# Proposal and Verification of Novel Machine Learning on Classification Problems

Chikako Dozono, Mina Aragaki, Hana Hebishima, Shin-ichi Inage

Fluid Engineering Laboratory, Department of Engineering, Fukuoka University, Fukuoka 8140180, Japan

**Abstract**

This paper aims at proposing a new machine learning for classification problems. The classification problem has a wide range of applications, and there are many approaches such as decision trees, neural networks, and Bayesian nets. In this paper, we focus on the action of neurons in the brain, especially the EPSP/IPSP cancellation between excitatory and inhibitory synapses, and propose a Machine Learning that does not belong to any conventional method. The feature is to consider one neuron and give it a multivariable $X_j$ (j = 1, 2,.) and its function value $F(X_j)$ as data to the input layer. The multivariable input layer and processing neuron are linked by two lines to each variable node. One line is called an EPSP edge, and the other is called an IPSP edge, and a parameter $\Delta_j$ common to each edge is introduced. The processing neuron is divided back and forth into two parts, and at the front side, a pulse having a width $2\Delta_j$ and a height 1 is defined around an input X . The latter half of the processing neuron defines a pulse having a width $2\Delta_j$ centered on the input $X_j$ and a height $F(X_j)$ based on a value obtained from the input layer of $F(X_j)$. This information is defined as belonging to group i. In other words, the group i has a width of $2\Delta_j$ centered on the input $X_j$, is defined in a region of height $F(X_j)$, and all outputs of $x_i$ within the variable range are $F(X_i)$. This group is learned and stored by a few minutes of the Teaching signals, and the output of the TEST signals is predicted by which group the TEST signals belongs to. The parameter $\Delta_j$ is optimized so that the accuracy of the prediction is maximized. The proposed method was applied to the flower species classification problem of Iris, the rank classification problem of used cars, and the ring classification problem of abalone, and the calculation was compared with the neural networks. As a result, we obtained more accurate results than neural networks in these classification problems.

## 1. Introduction

This paper aims to propose a new Machine Learning for classification problems. The classification problem is the process of classifying data into each category. Including pattern recognition, the classification problem is an important and popular problem in machine learning. In this paper, we consider the multiclass classification problem in particular. Machine learning approaches to classification problems include classification by Euclidean distance from a representative vector, neural networks, decision trees, Bayesian networks, clustering, and ensemble learning.

The classification by the Euclidean distance from the representative vector is considered to be able to represent each class by one representative vector, when the data of each class are locally aggregated for each class. This method can be further classified into Template Matching method [1], k-Nearest Neighbor method [2], etc.

Neural networks, especially hierarchical neural networks, are currently the mainstream classification method, and their development depends largely on (1) the ReLU function as an activation function of the intermediate layer [3], (2) the Softmax function as an activation function of the output layer, (3) the cross entropy function as a loss function [4], and (4) the stochastic gradient descent method as an optimization algorithm [5]. In particular, Stochastic Gradient Descent (Stochastic Gradient Descent: SGD) [6], AdaGrad [7], RMS prop [8], Adam [9] and many other reviews have been published. It is well known that many models have been proposed for the construction of neural networks that are specialized in various fields, such as convolutional neural networks, data compression by auto-encoders, and recurrent neural networks.

The decision tree [10] is a method of analyzing data using a tree structure (tree diagram). This method is particularly used in data mining because the process of classification in a classification model can be easily interpreted.

In this case, the decision tree has a tree structure in which the leaves represent the classification, and the branches represent the collection of features leading to the classification. The usefulness of the decision tree is that it is a nonparametric method that does not assume the distribution of the data to be analyzed. Both explanatory and objective variables can be used from nominal scales to interval scales and are said to be robust to outliers. On the other hand, classification accuracy is lower than other Machine Learnings, and it is not suitable for linear data.

The Bayesian network is one of the probabilistic inference techniques for probabilistic inference of events. By combining multiple relationships between cause and effect, the phenomenon that occurs while the cause and effect affect each other is visualized in terms of network diagrams and probabilities. The following features are available: 1) it is possible to analyze and infer the connection between "cause" and "effect," 2) when a certain "cause" is assumed, it is possible to infer the "effect" that may occur from it, and 3) when an expected "effect" is assumed, it is possible to infer the "cause" that may lead to it.

Clustering [11] and [12] are characterized by unsupervised learning, whereas neural networks classification is learning with supervised signals. K-means [13], an example of clustering, is a non-hierarchical clustering algorithm that uses the average of clusters to classify into a given number of k clusters. It is characterized by fast execution and scalability.

Thus, there are many approaches to the classification problem, and many faster and more accurate methods have been proposed.

The authors have proposed MOST (Monte Carlo Stochastic) [17], which is a new optimization method including learning of hierarchical neural networks, applied it to Iris classification problems, and verified its validity by comparing it with other optimization methods. In this paper, we propose an approach to a new classification problem based on the MOST optimization, which does not belong to any of the above, and verify its validity. In the modeling, we focused on the action of neurons in the brain, especially the EPSP/IPSP cancellation between excitatory and inhibitory synapses, and discussed it with reference to it. The details are described below.

## 2. Overview of neuron-especially synaptic action in the brain

The concept of the transmission mechanism between synapses is shown in FIG. 1 [14] and [15]. An axon from a nerve cell transmits a pulse of electrical potential, and a terminal synapse receives the pulse and releases a neurotransmitter. Through the synaptic cleft, neurotransmitters reach the postsynaptic membrane of the dendrites of neighboring neurons. There are two types of synapses associated with neurons in the brain: excitatory and inhibitory. As shown in FIG. 2 a), an excitatory synapse stimulates sodium receptors on the postsynaptic membrane by releasing excitatory transmitters, such as glutamate, while maintaining a constant gap with the postsynaptic membrane of the next neuron's dendrite. The receptors transmit $Na^+$ ions and generate an excitatory postsynaptic potential (EPSP) in the neuron. Conversely, the inhibitory synapses shown in FIG. 2 b) stimulate chloride receptors on the postsynaptic membrane by releasing inhibitory transmitters such as GAVA and glycine. Receptors transmit $Cl^-$ ions to generate inhibitory postsynaptic potentials (IPSP) in the neuron. Interference between EPSP and IPSP results in time-weighting, space-weighting, and EPSP/IPSP cancellation as shown in Table 1. The time weighting in Table 1 is the transmission of multiple EPSPs with time differences from one synapse, and the superposition of these EPSPs results in EPSPs with larger potentials. The spatial weights in Table 1 represent the transmission of a single pulse of EPSP from multiple excitatory synapses, which are superimposed. Finally, EPSP/IPSP cancellation results from the superposition of opposite potentials at excitatory and inhibitory synapses, resulting in inhibition. These three actions are considered to provide appropriate signal transduction processing in the brain. In this paper, we propose an algorithm for a new classification problem, taking a hint from this structure consisting of excitatory and inhibitory synapses, and especially from EPSP/IPSP cancellation.

## 3. Proposed Model

### 3.1 Case of single variable：

In the previous chapter, we described the nature of synapses in neurons. The system of the proposed model using a single variable function as an example is shown in FIG. 3. First, consider multiple (x, F(x)) data sets-Teaching signals. Select element i: ($X_i$, F($X_i$)) as one of the elements. The proposed system does not require complex networks like conventional neural networks. The input node is given a variable $X_i$ with a certain value and its output F($X_i$). The input node and operation node of variable $X_i$ are connected by two red and blue lines, and the red line is named EPSP edge and the blue line IPSP edge. The processing at each edge is described below with reference to FIG. 4. Each edge is given a parameter Δ with a common value. At the EPSP edge shown in FIG. 4 a), the well-known step function U (x) is shifted by ($X_i - \Delta$). That is, the following.

$$U(x) \quad \rightarrow \quad U\big(x - (X_i - \Delta)\big) \tag{1)}$$

were,

$$U(x) \begin{cases} = 1 & (0 \leqq x) \\ =0 & (0 > x). \end{cases} \tag{2)}$$

In the figure, the red line indicates the step function, and the yellow-green line indicates the function after processing.

Next, at the IPSP edge shown in Fig. 4 b), the well-known step function multiplied by -1 is shifted by ($X_i + \Delta$). In other words, do the following.

$$-U(x) \quad \rightarrow \quad -U\big(x - (X_i + \Delta)\big) \tag{3)}$$

In the figure, as in the case of the EPSP edge, the red line indicates the step function, and the yellow-green line indicates the step function on the negative side after processing.

The operation node in FIG. 2 is divided into two parts, the front part and the back part, and the following ψ is calculated in the front part.

$$\psi(X_i) = U\big(x - (X_i - \Delta)\big) - U\big(x - (X_i + \Delta)\big) \tag{4)}$$

This treatment corresponds to the EPSP/IPSP cancellation described in the previous chapter. As shown in FIG. 4c), ψ is a pulse function of width 2Δ and height 1.0 given by Eq. 4). Finally, at the back of the node, the following φ is calculated by multiplying the other input F($X_i$) by ψ.

$$\varphi(X_i) = F(X_i) \times \psi \tag{5)}$$

The range defined from $X_i$ and F($X_i$) is called Group i. The meaning is simple: "For a variable x in the range $X_i-\Delta$ to $X_i+\Delta$, all outputs are F($X_i$)."

Next, consider a case where this processing is applied to multiple data items using FIG. 5. The dashed red line in the figure is the function behind the data. As shown in FIG. 5 a), ($X_1$, F ($X_1$)) determines Group-1 under a certain Δ. These operations are called learning and memory. Next, if the above processing is performed with new data, such as $X_2$ and $X_3$, then Groups-2 and -3 can be obtained. This repetition is called experience. Next, consider new data represented by red and green triangles as shown in FIG. 5c). Because the data represented by the red triangle belongs to Group-2 in the figure, the output is expected to be $F(X_2)$. On the other hand, the data represented by the green triangle does not belong to any group, so Group-4 based on the new green triangle data is learned. The processes in FIG. 5 d) are called "relearning" and "additional memory." By repeating this operation, you can learn by the Teaching signals, make predictions in the memorized group, and if there is no group to which you belong, you can construct a smarter group one after another by repeating relearning. The above is an outline of the proposed model. Although our proposal has a simple structure as shown in FIG. 3, it can express processing similar to the process that our brain learnings. If this operation is applied to a continuous discretized variable at a constant interval $\Delta$, the function F(x) can be approximated depending on the resolution of the given $\Delta$, as shown in FIG. 6. It is easy to imagine that the discretization function asymptotically approaches the original function F(x) at $\Delta \rightarrow 0$. Presumably, this function approximation with continuous discretized variables at constant intervals $\Delta$ is equivalent to the universal function approximation theorem [16], which is the theoretical validity of neural networks. Learning in this model is to optimize $\Delta$ associated with EPSP and IPSP edges. This optimization of $\Delta$ is described in detail in the algorithm in the next chapter.

### 3.2 Case of multivariable：

Again, the basic idea is the same as for a single variable. The case of two variables is shown in FIG. 7. The input nodes are $X_{1i}$, $X_{2i}$ and $F(X_{1i}, X_{2i})$. The $X_{1i}$ node, the $X_{2i}$ node, and the arithmetic node are connected by two lines of EPSP edge and IPSP edge, respectively, as in the case of a single variable. Processing at each node and edge is summarized below.

1) Processing on EPSP Edge:

$$U_j(x_j) \quad \rightarrow \quad U_j\left(x_j - \left(X_{ji} - \Delta_j\right)\right) \qquad (j=1,2) \tag{6)}$$

2) Processing on IPSP Edge:

$$-U_j(x_j) \;\rightarrow\; -U_j\left(x_j - \left(X_{ji} + \Delta_j\right)\right) \qquad (j=1,2) \tag{7)}$$

3) Processing of $\psi$:

$$\psi(X_{1i}, X_{2i}) = \prod_j \left[U_j\left(x_j - \left(X_{ji} - \Delta_j\right)\right) - U_j\left(x_j - \left(X_{ji} + \Delta_j\right)\right)\right] \tag{8)}$$

4) Processing of $\varphi$:

$$\varphi(X_{1i}, X_{2i}) = F\left(X_{1j}, X_{2j}\right) \times \psi(X_{1i}, X_{2i}) \tag{9)}$$

It differs from a single variable only in that $\psi$ is the product of the pulse functions of variables $X_1$ and $X_2$. The range defined here, and its output $F(X_{1i}, X_{2i})$ are called group-i. As in the case of a single variable, the meaning is that the outputs of variables $x_1$ and $x_2$ in the range: $X_{ji}-\Delta_j$ ─ $X_{ji}+\Delta_j$ (j = 1, 2) are all $F(X_{1i}, X_{2i})$. The "learning" and "memory" of these two variables are shown in FIG. 8. Other operations in FIG. 5 ─ operations such as "experience," "prediction," and "relearning" are the same as for a single variable. This concept can be applied to any number of variables, and only the product of the pulse function increases as the number of variables increases in Eq. 8). Learning in the case of multiple variables also involves optimizing $\Delta_j$ associated with the EPSP edge and IPSP edge of each variable. The systems shown in FIGs. 3 and 7 are similar to the structure of a perceptron, but the value of the function F(x) is not an output but an input, and the processing at the operation node is completely different. The number of variables to be optimized coincides with the number of inputs parameters, and is much less than that of conventional neural networks, which is expected to reduce the computational load. In addition, it is well known that when a Teaching signal are biased to a specific data, it is strongly affected by data with many learning results of conventional neural networks. However, the proposed algorithm has a feature that it is not affected by the amount of data because the weights of the groups obtained are the same. Hereinafter, this model, which is common to both single and multiple variables, is called Single Neural Grouping (SiNG) Method.

## 4. Algorithm of proposed model

### 4.1 Fundamental algorithm

The object of this paper is multiclass classification among classification problems. For example, in the Iris classification problem described below, flowers are classified into three types based on the characteristics of petal length and width and calyx length and width data, and handwritten numbers are classified into 10 numbers of 0－9, etc. For this reason, in multi-class classification, the input of multiple variables is a numerical value representing the category in which the output is classified.

Consider a case where there are N Teaching signals (N is an even number) for M variables ($x_1$, $x_2$, $x_3$,..., $x_M$) and their function F($x_1$, $x_2$, $x_3$,..., $x_M$). The flow of the algorithm is shown below.

[STEP -1] For each variable ($x_1$, $x_2$, $x_3$,..., $x_M$), an appropriate initial value $\Delta_j$ (j = 1, 2..., M) is given.

[STEP -2] The Teaching signals are divided into two groups, the other being Data group-I and the other being Data group-II.

[STEP -3] Define the range of group i with each data i: ($X_{1i}$, $X_{2i}$, $X_{3i}$,..., $X_{Mi}$) belonging to Data group-I. That is, the range of each variable in group i: [$X_{ji}$-$\Delta_j$, $X_{ji}$-$\Delta_j$] is calculated. The output of the variables within the range ($x_1$, $x_2$, $x_3$,..., $x_M$) is the same at F($X_1$, $X_2$, $X_3$,..., $X_M$). Since N/2 data belonging to Data group-I is used for grouping, the number of groups is N/2.

[STEP -4] Check whether each data from Data group-II belongs to one of the N/2 groups determined in [STEP-3]. Now, when the TEST signals is ($X_{1T}$, $X_{2T}$, $X_{3T}$,..., $X_{MT}$), the number of variables $X_{jT}$ (j = 1,... M) satisfying the range of each variable in a certain group i: $X_{ji} - \Delta_j, < X_{jT} < X_{ji} + \Delta_j$ is counted, and if the total number of these variables is M, then we belong to the group.

[STEP -5] In [STEP -4], if ($X_{1T}$, $X_{2T}$, $X_{3T}$,..., $X_{MT}$) belongs to a single group, the output value of that group is used as the output of the TEST signals. On the other hand, if the positions of the Teaching signals are dense and $\Delta$ is large, the TEST signals may belong to multiple groups, as shown in FIG 9. FIG. 9 (a) shows that the TEST signals belong to four groups when it belongs to multiple groups with the same output, and b) shows that it belongs to groups with different outputs, one group for output A and three groups for output B. In the proposed algorithm, consider a case where the output overlaps the groups A and B as shown in Fig. 9b). Expected output value - for example, 0－9 for numeric recognition - counts the number of groups

that belong to each. The output of the group with the highest count P(k) (k: output value) is used as the final TEST signals output.

**[STEP-6]** In the processes [STEP-1] - [STEP-5], as shown in the green triangle in FIG. 5 c), some data do not belong to any of the groups determined from Data group-I. In this case, the output should contain values that are different from those you want to classify. In addition, we add this case as a new group as "relearning " result.

**[STEP-7]** In the above steps [STEP-1] - [STEP-6], the value of $\Delta_j$ for each variable is optimized to maximize the accuracy of prediction of the output of the TEST signals. The optimization method is described in detail in the next chapter.

**[STEP-8]** In addition, the Data groups-I and II defined in [STEP-2] are replaced, and the value of $\Delta_j$ is optimized in [STEP-1] - [STEP-7] above, with data group II as the Teaching signals and data group I as the TEST signals.

**[STEP-9]** Compare $\Delta_j$ predicted from Data group-I → Data group-II with $\Delta_j$ predicted from Data group -I → Data group-II, and use the larger one as the final $\Delta_j$.

**[STEP-10]** In the above flow, the whole Teaching signals are divided into a new Teaching and TEST signals to optimize $\Delta_j$, so in this STEP, the prediction is based on a separately prepared TEST signals. Since the optimization of $\Delta_j$ has already been determined, the above steps [STEP-1] - [STEP-6] are simply applied to the TEST signals.

This is the algorithm of the proposed model. The overall flow of the algorithm is shown in FIG. 10.

### 4.2 Optimization method－MOST(Monte Carlo Stochastic) method

In conventional neural networks, the weight coefficients defined on the edges connecting nodes are optimized based on the error back propagation method, using algorithms such as Adam, RMSprop, and Adamax. In the proposed algorithm, we apply an optimization method called MOST that we have developed separately. MOST always generates a convergent solution, and has features such as high speed and high accuracy compared with Genetic Algorithm (GA). MOST has also been used for weight optimization of conventional neural networks and has been tested [17].

The proposed Monte Carlo optimization method is based on the following propositions. "In an objective function defined by a finite interval and having an extreme value of a finite number M, we consider dividing a prior-period defined interval by a number N sufficiently larger than M. When the objective function is integrated in each division interval, the minimum value of the objective function exists in the interval with the smallest integration value." Assuming this, the following optimization algorithm is considered. We now consider an objective function with multiple extremes, as shown by the red line in the upper left of FIG. 11. This function is defined in the range [0, 2.5]. Consider the optimization to find the smallest inflection point in this region. In the case of such an objective function that waits for multiple inflection points, a local solution is often obtained, and the correct optimization cannot be performed.

**[STEP-1]** Divide the variable parameter domain by a large number, such as 20. The integration is performed in each divided region. By comparing each integral value, the division region which takes the smallest integral

value is selected from them. In FIG. 11, the upper part and the second figure, the integration values enclosed in pink are the minimum. It is expected that there are variable parameters which minimize the objective function in this region [1.375, 1.5]. And it can also be expected that there is at most one extreme value in the division interval by dividing with sufficiently large number. This operation prevents falling into the local solution. In this method, the integration is calculated by numerical integration, especially the Monte Carlo method.

**[STEP-2]** The section (In the case of FIG. 11, the area of [1.375, 1.5]) of the variable parameter selected in STEP -1 is divided into two. Since it is expected that there is at most one extreme value in this interval, compare the integrated values in the two divided intervals, and select the interval with the smaller integration value again.

**[STEP-3]** In STEP-3, the process of STEP-2 is repeated to compare two divisions and integral values until the width of the division section becomes $< 10^{-6}$. The average of the minimum and maximum values of the interval at that time is a combination of multivariable parameters that minimizes the desired objective function. When the bisection is repeated K times for a region, the width of the region decreases at $(1/2)^K$. For example, if the defined width is 1, and you want to converge to $< 10^{-6}$, then converge at $(1/2)^K < 10^{-6}$ is established. I.E. K.> 20.

This approach to optimization is called MOST. The flow of each step of MOST is shown in FIG. 10. If the number of extremes of the objective function is sufficiently small, or if the minimum and other extremes are sufficiently different, the correct solution may be obtained even if STEP-1 is omitted. Next, we extend the method to an objective function consisting of multiple variables. In the case of multivariables, the flow shown in FIG. 10 can be applied as is. The Monte Carlo integration of the multivariable objective function $f(x_1, x_2, x_3, ..., x_n)$ can be calculated as follows using uniform random numbers in each region $[a_n, b_n]$ (n = 1, 2,..., M) of multivariables: $x_1, x_2, x_3, ..., x_n$.

$$\text{Numerical integration} = \frac{1}{P}\frac{1}{(b_1-a_1)}\frac{1}{(b_2-a_2)}\cdots\frac{1}{(b_M-a_M)}\sum_{j=1}^{K} f(X_{1j}, X_{2j}, \cdots X_{nj}) \qquad 10)$$

where n is the number of variables and K is the total number of Monte Carlo calculations. The Monte Carlo method generates random numbers corresponding to each variable even in the function of multivariable, substitutes them to the objective function, adds them up, and divides them by the number of random numbers to obtain the numerical integration value. Thus, the optimization method based on the Monte Carlo method can be applied to an objective function composed of multiple variables. However, in the case of multivariable, when each is divided into two at once, the number of divisions of the region becomes $2^n$. For example, when n is 100, the number of divisions is $1.26 \times 10^{30}$. And, the real calculation becomes difficult, when the number n of the variable increases, because the integration is repeated 20 times necessary for the solution to converge, and as a result, it integrates $2^n \times 20$ times in total. To solve this problem, we first divide only the variable $x_1$ into two and fix the domain of the other variables. The integration value is calculated by the Monte Carlo method, and the region in which the integration value is small is chosen. Next, one side region of selected $x_1$ is fixed, and only the region of $x_2$ is divided into two. The domain of variables after $x_3$ is fixed. In addition, the integration value calculation and the region of $x_2$ are selected again by the Monte Carlo method. By repeating this process with other variables xi, all variables converge as optimal solutions.

In this case, the number of divisions is $2 \times n$ even if there are n variables, and $2 \times n \times 20 = 40 \cdot n$ even if 20 times are required for convergence. Therefore, even if n is 100, the number of calculations is only 4000, and the number of calculations can be drastically reduced without exponential increase. The comparison is shown in FIG. 11. To apply the proposed model described above, the error between the output and the predicted value in the TEST signals, which is defined below, is minimized instead of the integral value to obtain the optimum $\Delta_j$.

$$\text{Error} = \frac{1}{2}\sum(\text{Truth value} - \text{Predicted value})^2 \qquad 11)$$

The smaller the error, the higher the accuracy rate. For this loss function, there is an option to use cross entropy or the like.

## 5. Verification of Proposed Model with Actual Classification Problem

The proposed method combined with MOST was applied to three problems: (1) Iris flower species classification problem, (2) used car rank assessment problem, and (3) abalone age classification. These three classification problems are sufficiently reliable as validation benchmark problems for neural networks.

### 5.1 Neural networks conditions used for comparison

In this paper, we calculated the classification problems of Iris, used cars, and abalone by using a network with two hidden layers between the input layer and the output layer. The LeRu function is applied to the first hidden layer and the SoftMax function is applied to the second hidden layer. The last output layer selects the one with the larger SoftMax function value as the final output result. From the difference between this output layer and the flower species number of the actual learning data, apply the same squared error as in Eq. 11). To minimize the total error of the learning data, the weighting factor on the line connecting each node and the bias node is optimized. The MOST described above is used for optimization. The bias is applied to the input layer and the hidden layer, respectively.

### 5.2 Iris floral classification problem

Data on four parameters of Iris, namely, “Sepal length”, “Sepal width”, “Petal length” and “Petal width”, are given to classify Iris into three flower varieties, “Versicolor”, “Setosa” and “Virginica”, depending on their characteristics [18] and [19]. To each flower species give the numbers Versicolor: 1, Setosa: 2, and Virginica: 3. The comparison between the proposed method and neural networks in this Iris classification is shown in Figure 13. The hidden layers -1 and 2 of the neural network are each provided with 3 nodes. Including bias, the neural networks have 27 weighting factors to determine. The proposed method requires only $\Delta_1$ to $\Delta_4$ to be determined for each of inputs $X_1$ to $X_4$, and the number of variables to be optimized is smaller than that of a neural network. This feature reduces the computational load in optimization. Inputs $X_1$-$X_4$ are given by four input data obtained from the UC Irvine Machine Learning Repository: “Sepal length”, “Sepal width”, “Petal length”, and “Petal width”. Examples are shown in Table 2. The 150 data consist of “Versicolor”:50 data, “Setosa”:50 data, and “Virginica”:50 data, of which 120 data are used for training as Teaching signals and the remaining 30 data are used for testing. In the TEST signals, 10 data were randomly selected from the original 150 data for each flower species. Optimization in MOST requires specifying the search area and the number of random numbers to be used for Monte Carlo integration. For SiNG, the search region is $0<\Delta_j<1.0$ and the number of random numbers is 50. For neural networks, the search region is - 2.0<$w_{ij}$<2.0 and the number of random numbers is 200.

Tables 3 and 4 show the results of optimizing the weighting coefficient of the neural networks: $w_{ij}$ and the width of the SiNG: $\Delta_j$. Tables 3 and 4 show the results of optimization of the neural networks: $w_{ij}$ and SiNG: $\Delta_j$. Table 5 shows the actual flower species and SiNG evaluation results for the 30 TEST signals using the optimized values. The 3 columns to the right of the table are the count variables P (k) of the group to which they belong, as shown in [STEP -5] of the basic algorithm. Each TEST signal may belong to as many as 40 groups. However, in the range of this calculation, it can be judged that the appropriate optimization was carried out, because there was no case which spanned the groups of different outputs. In the case of applying SiNG and neural networks to 30 TEST signals, the predictive accuracy rates are compared in Table 6. The neural networks have good prediction accuracy with 99% correct rate in learning and 93% correct rate in prediction. The results obtained by SiNG were 100% for both learning and prediction, and it was confirmed that more accurate results were obtained. It is a well known fact that in neural networks, especially when the number of nodes is small, learning results differ for each optimization, and learning must be repeated to increase the accuracy rate. On the other hand, in SiNG, the reproducibility of the learning of Δj is high and the repetition of the learning is not required.

### 5.3 Rating classification problem of used cars

Based on six parameters, “Buying Price”, “Price of Maintenance”, “Doors”, “Persons Capacity”, “Size of luggage boot”, and “Safety of the car”, used car rating is classified into four categories: “Unacceptable”, “Acceptable”, “Good”, and “Very Good”. Assign each of the 4 ranks a number of 1－4 [20], [21]. FIG. 14 compares the SiNG and neural networks for vehicle classification. In the neural networks, there are 2 hidden layers, each with 10 nodes and 8 nodes, and the number of weights to be optimized is 194. As described above, there are 4 outputs, “Unacceptable”, “Acceptable”, “Good”, and “Very Good”, each with a rank of 1－4. As input data, 995 learning data and 729 TEST signals were quoted from Irvine Machine Learning Repository. As an example, a part of the data of the Teaching signals are shown in Table 7. The table gives the numerical values for each parameter, and Table 8 shows how to determine the numerical values. For example, "Buying Price" is given as “Low”, “Medium”, “High”, and “Very High”, and 1－4 is given to quantify it. As with Iris, optimization in MOST requires specifying the search area and the number of random numbers to be used for Monte Carlo integration. For SiNG, the search region was set to 0<$\Delta_j$<5.0 and the number of random numbers was set to 50. For neural networks, the search region was set to -2.0 <$w_{ij}$<2.0 and the number of random numbers was set to 50.

The optimized values of $\Delta_j$ for SiNG are shown in Table 9. The optimized values are applied to predict the output of 729 TEST signals, and the accuracy is compared with Table 10. The neural networks have a fixed prediction accuracy with 84% correct rate in learning and 81% correct rate in prediction. On the other hand, the results obtained by SiNG were 100% for learning and 94% for prediction, and it was confirmed that the results obtained by SiNG were more accurate than the results obtained by neural networks.

### 5.4 Abalone ring classification problem

The annual rings of abalone are related to 8 parameters: “Sex”, “Length”, “Diameter”, “Height”, “Whole Weight”, “Shucked Weight”, “Viscera Weight”, and “Shell Weight” [22] and [23]. The distribution of annual rings is between 1 and 29. In this case, we refer to the result of Egemen Sahin et al. as a calculation of neural networks. As input data, 2000 pieces of learning data and 2117 pieces of TEST signals were quoted from Irvine Machine Learning Repository. Table 11 shows examples of Teaching signals. As described above, rings are widely distributed up to 1 －29, and it is difficult to predict them directly. Therefore, Egemen Sahin et al. treated the rank of rings by classifying them into 3 categories: < 9, 9 ≦ rings < 18, and 18 or more. They also used a feedforward back propagation artificial neural network network (FFBANN), a convolutional neural network (CNN), and a residual neural network to improve classification prediction accuracy. The neural network consists of 2 hidden layers with 65 and 65 nodes, respectively. As a result, 79% of the TEST signals were correct. The model configurations of both neural networks and SiNG are the same as those of a) in FIGS. 13 and 14, and are therefore omitted. Table 11 summarizes the optimization results of $\Delta_j$, and Table 12 compares the results of SiNG and neural network. The accuracy of SiNG was 85%, which was better than that of the neural network.

The validity of the proposed method was verified by applying SiNG to the above three cases and comparing with the results of neural networks.

## 6． Conclusions

We focused on the action of neurons, especially excitatory synapses and inhibitory synapses, and proposed a new approach to classification problem: SiNG. The main features and algorithms are as follows.

1） Variable: xi is sent to the operation node through the EPSP edge and IPSP edge to generate a pulse of width 2Δ and height 1.0. Multiply by the input $F(x_i)$ to generate $F(x_i)$ pulses of width 2Δ and height. An area with this information is called a group. M groups are defined by processing only M Teaching signals.

2） Compare the TEST signals with the group obtained in 1), and use the pulse height of the group to which it belongs as the output of the TEST signals.

3） Based on this concept, Δ is optimized so as to maximize the accuracy rate that the predicted value of the TEST signals matches the true output.

This method was applied to Iris, used cars and abalone classification problems. It was confirmed that the accuracy rate was 85－100%, which was higher than that of conventional neural network. This proposed method is considered to be unique and does not belong to any of the conventional classification problem algorithms. It is well known that when a Teaching signal is biased to a specific data, it is strongly affected by the data with many learning results of conventional neural networks. However, in the proposed algorithm, the weights of the obtained groups are the same, and it is not affected by the amount of data.

In addition to the present verification, we plan to further verify the pattern recognition by applying it to image data, etc. and to study an extension method other than the classification problem.

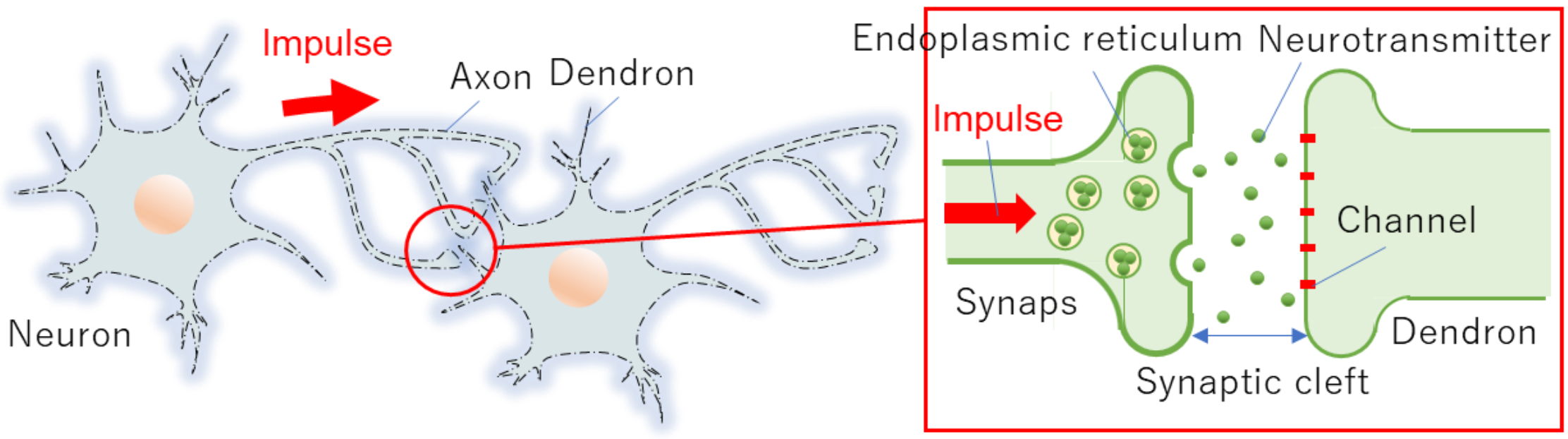

FIG. 1: Synaptic Transmission

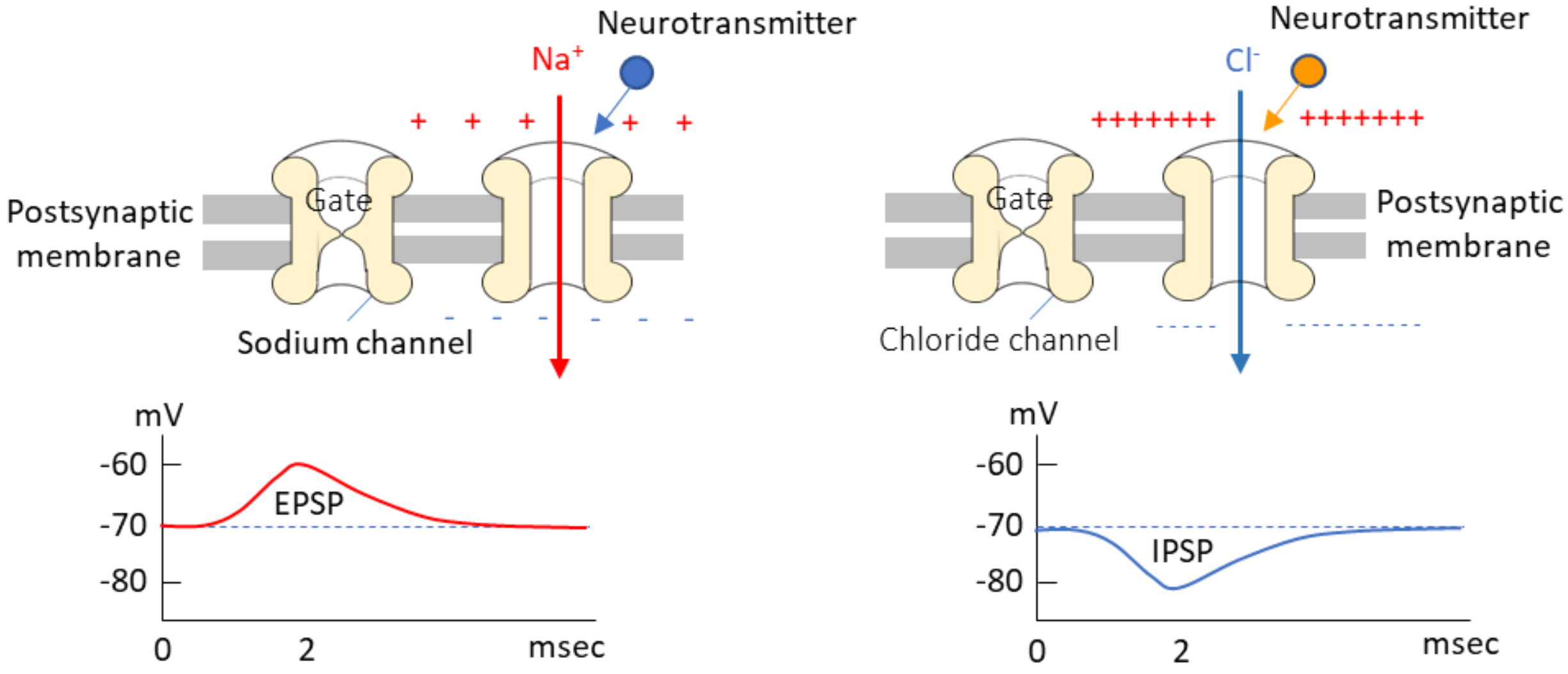

a) Excitatory synapse b) Inhibitory synapse

FIG.2: Actions of Excitatory and Inhibitory synapses

Table.1: Sequence of spatial and temporal summation of EPSP and IPSP. EPSP-IPSP cancellation

| Temporal summations | Spatial summations | EPSP-IPSP Cancellation |
|---|---|---|
| 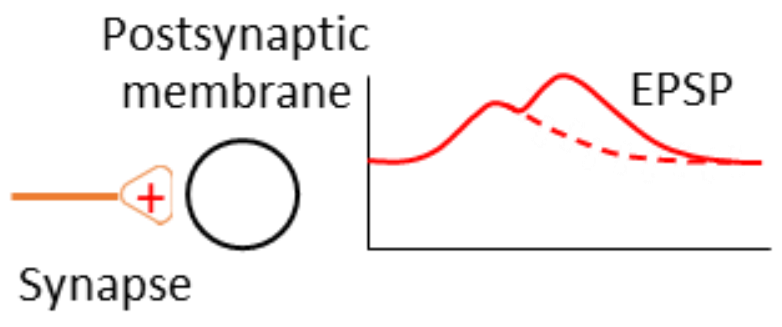  | 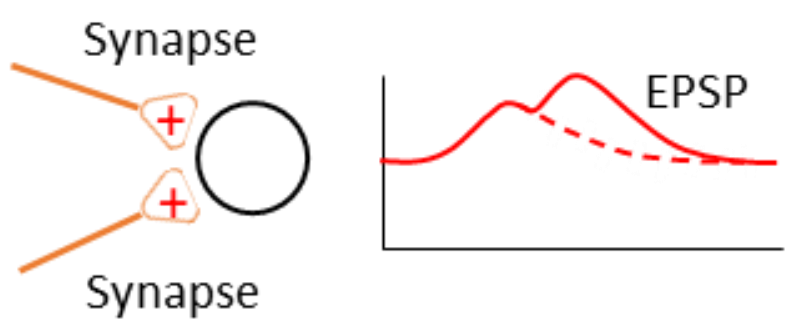  | 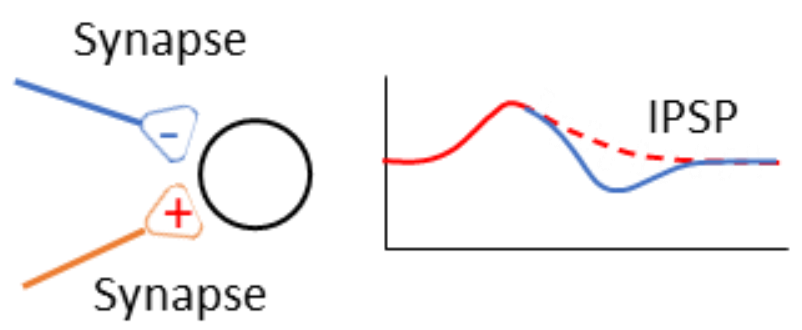  |

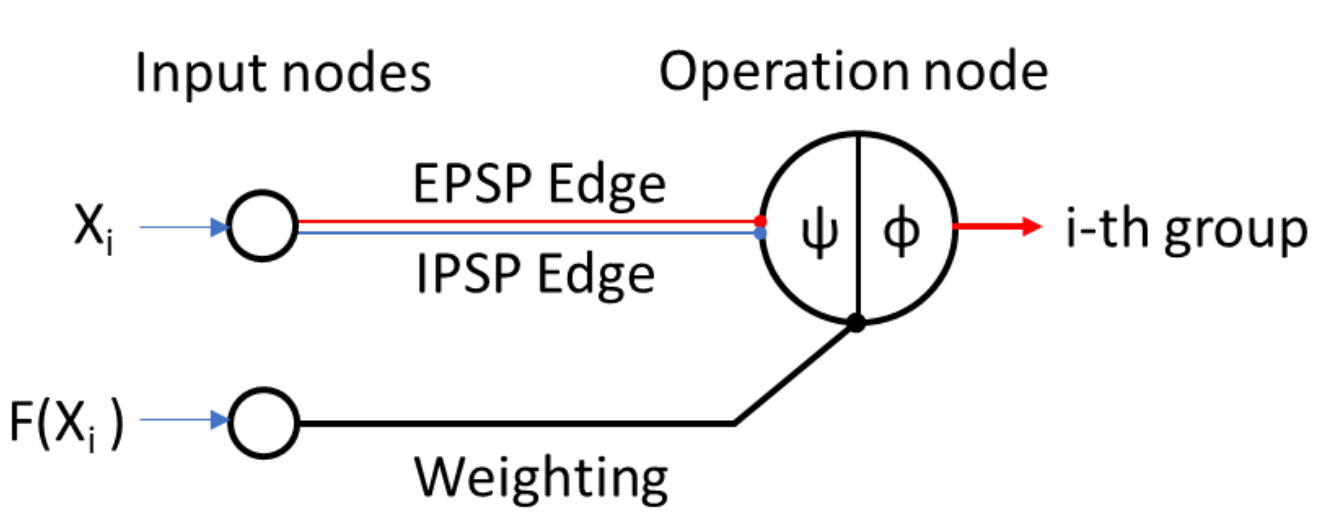

FIG. 3 : Conceptual configuration of SiNG

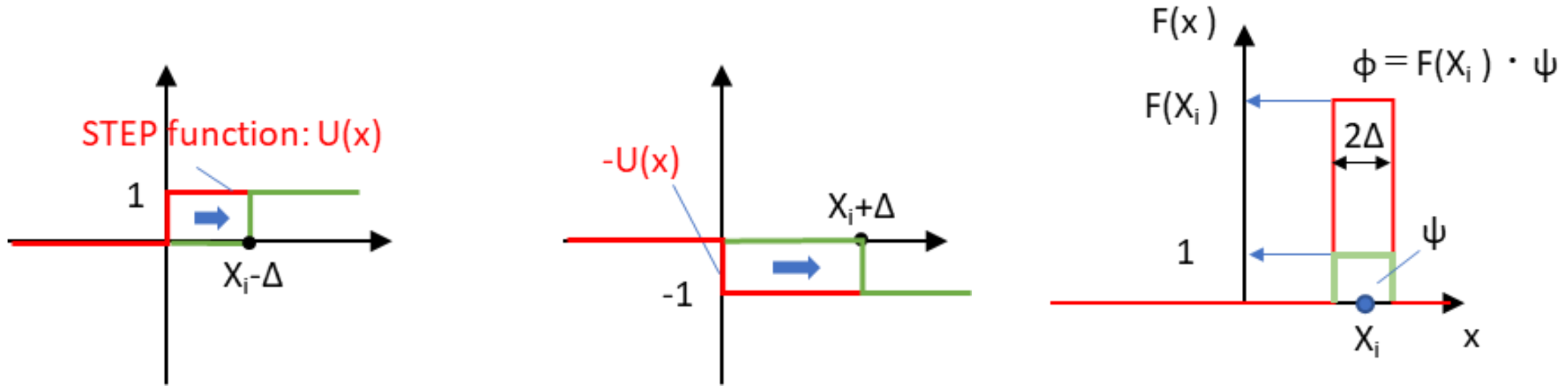

a) On the EPSP edge b) On the IPSP edge c) On the operation node

FIG. 4: Operation of each part of the system

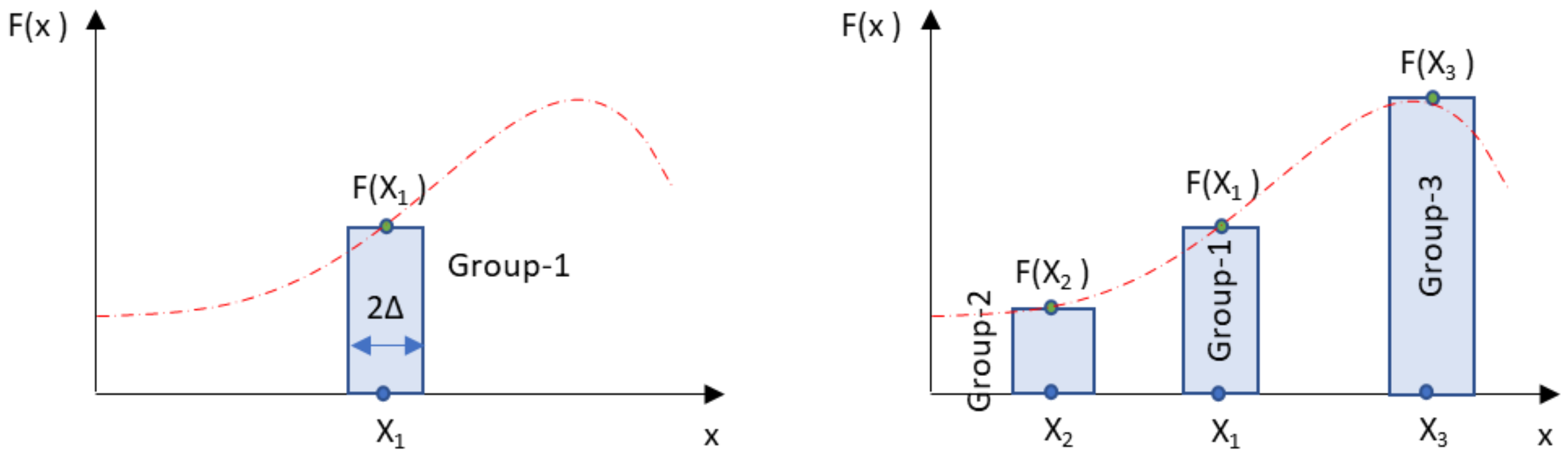

a) Learning and memory operations b) Experience operations

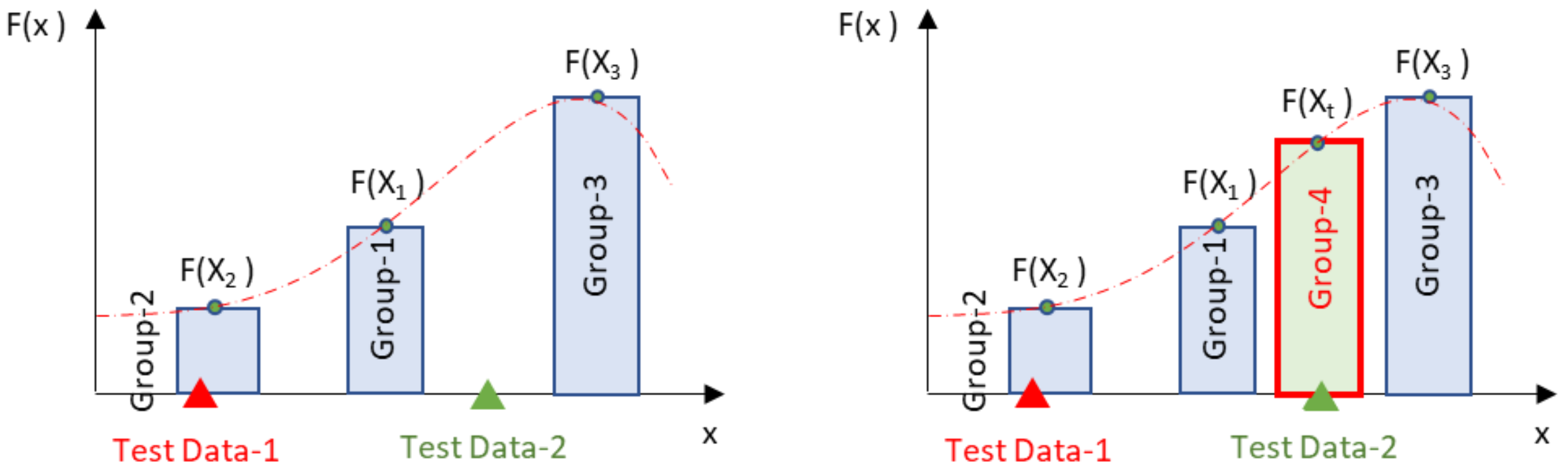

c) Prediction operation d) Relearning and additional memory operations

FIG. 5: Fundamental operations on SiNG

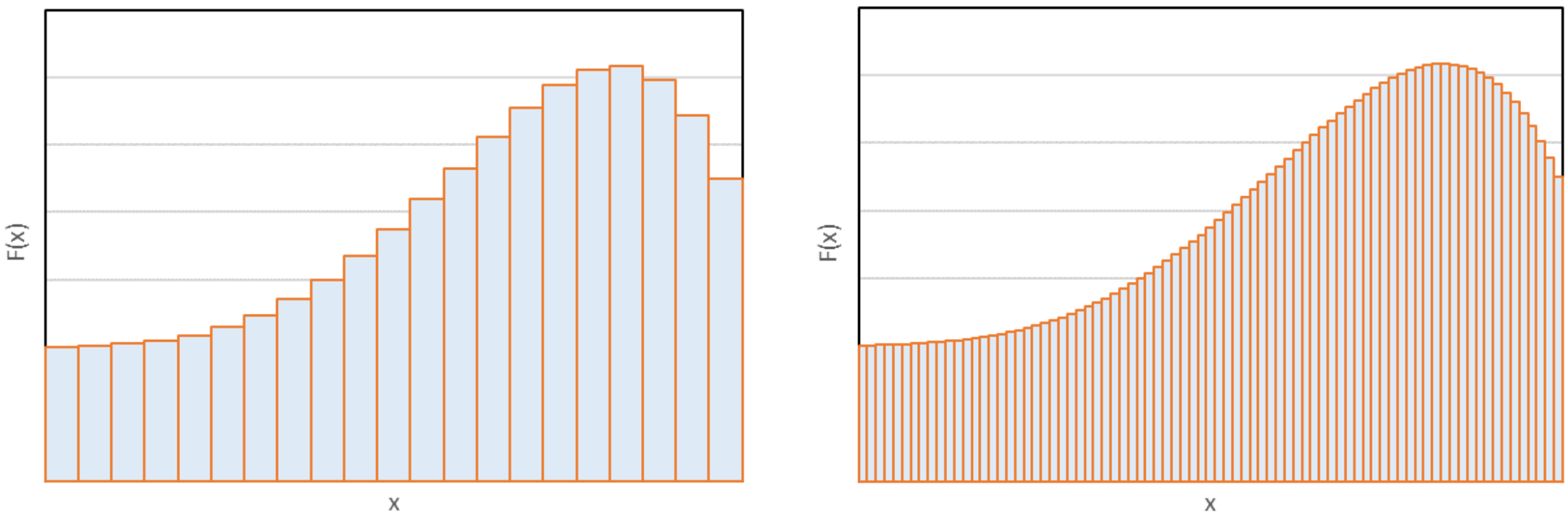

**a)** Coarse approximation by continuous discretization b) Lower Δ for continuous discretization

FIG. 6: Continuous Discretization Function Approximation by SiNG

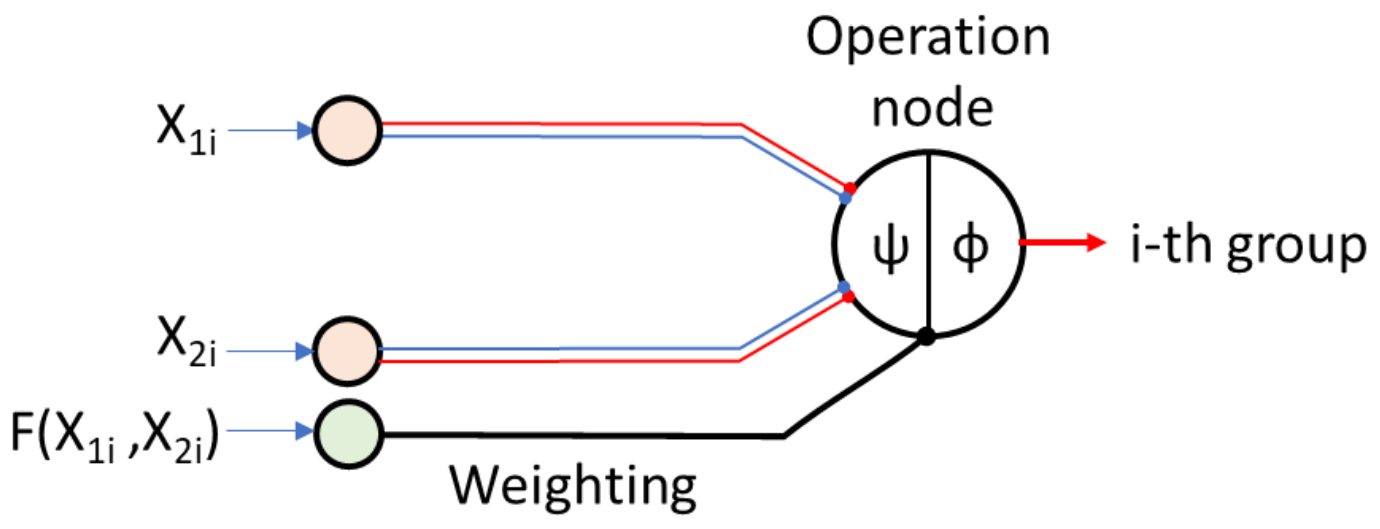

FIG.7: Basic configuration of bivariate SiNG

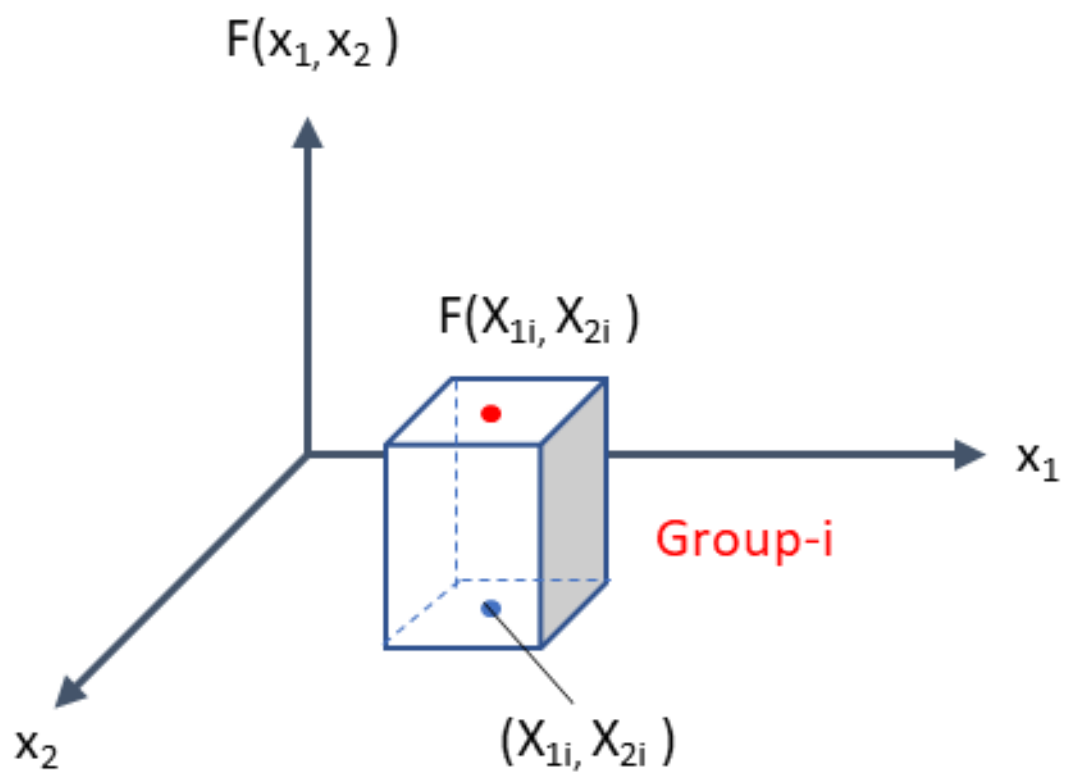

FIG. 8: Operation on the operation node

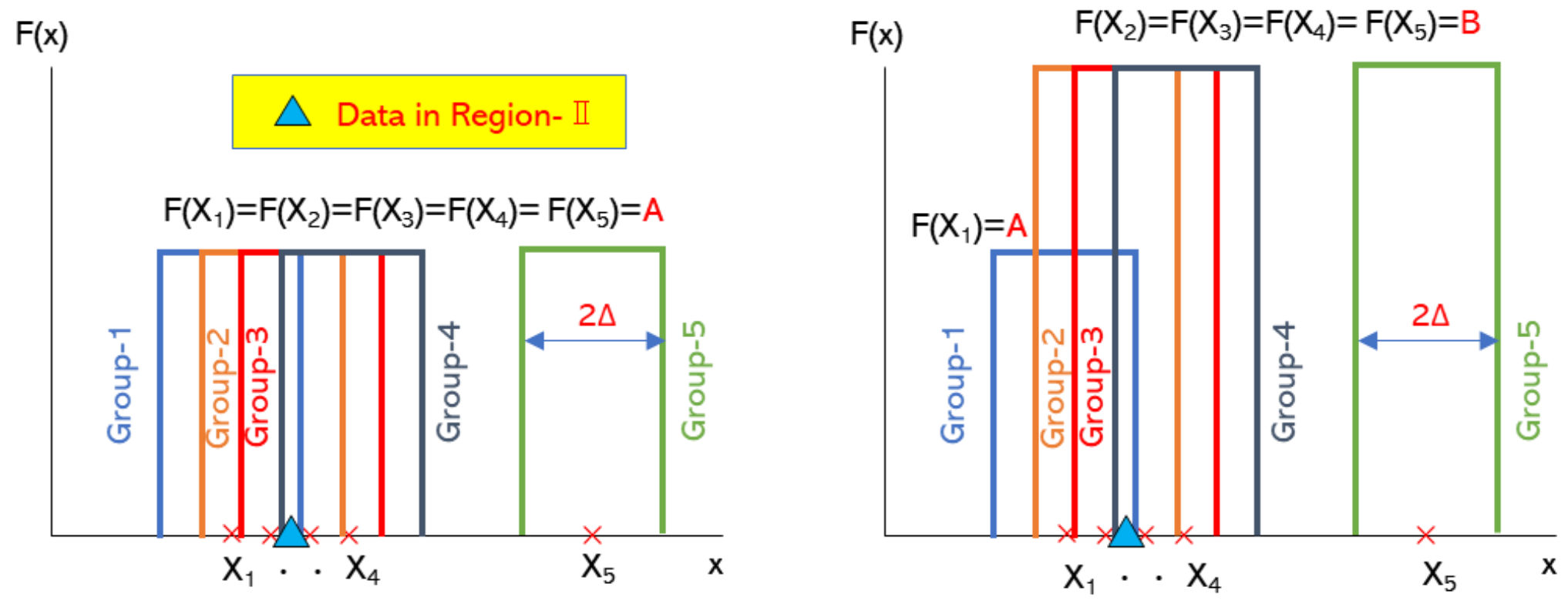

a) Duplication in groups of the same output b) Duplicate in groups of different outputs

FIG.9: Example of Processing in overlapping groups

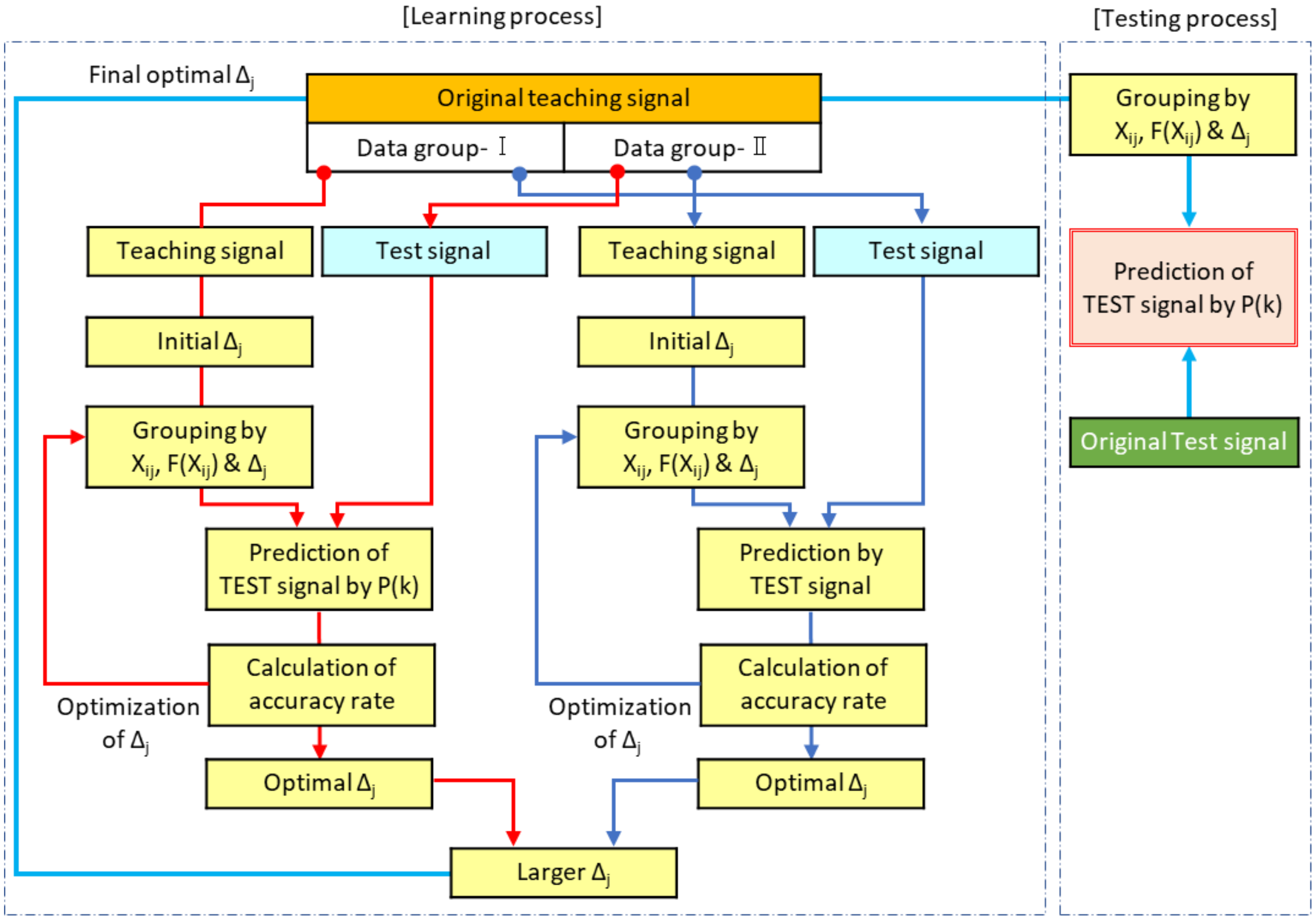

FIG. 10: Overview of SiNG algorithm

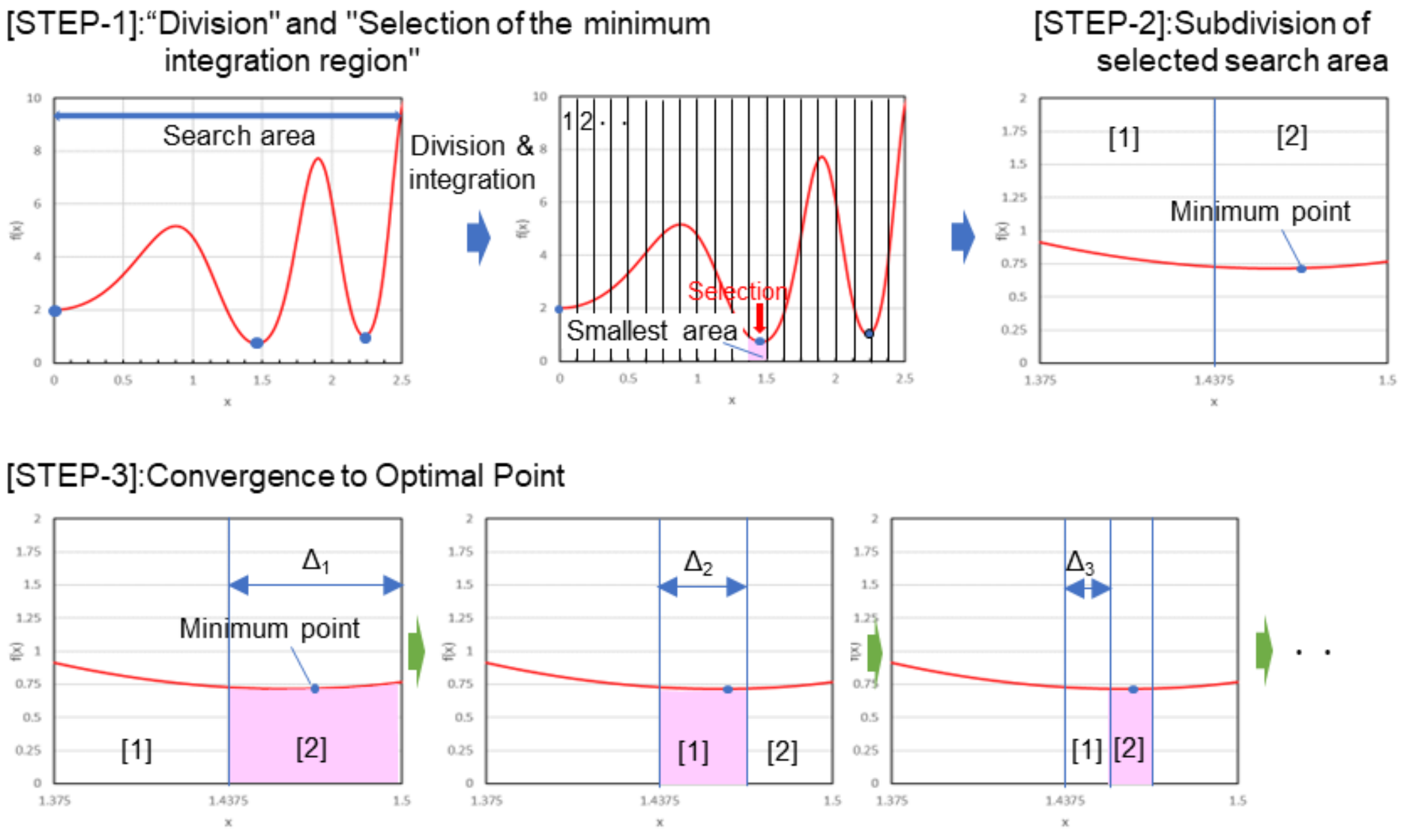

FIG. 11: Flow of proposed optimization-MOST

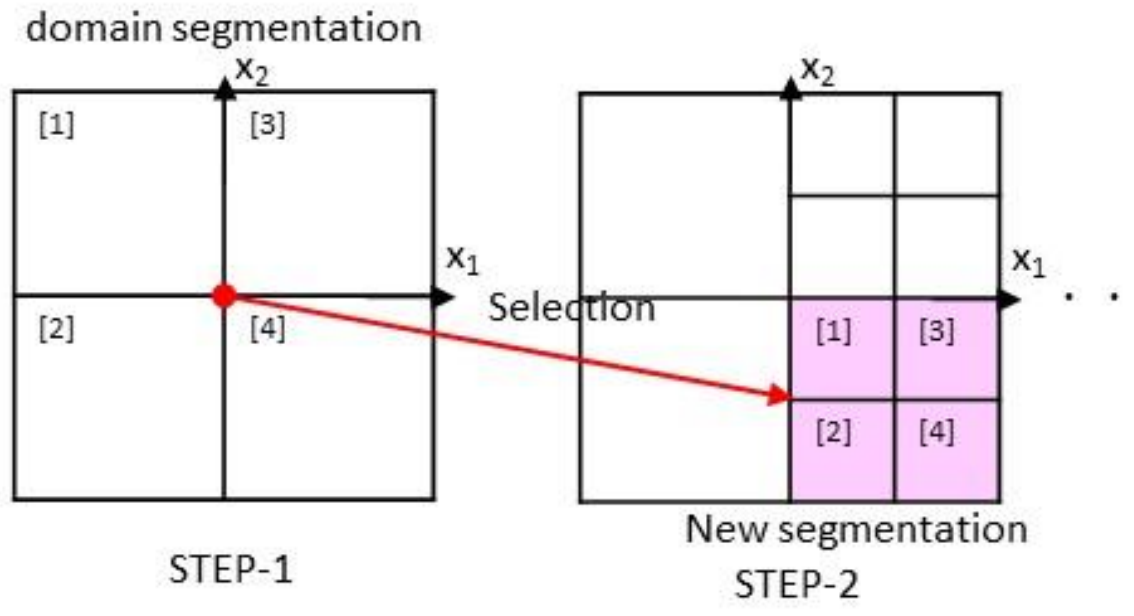

1) Equally divided segmentation

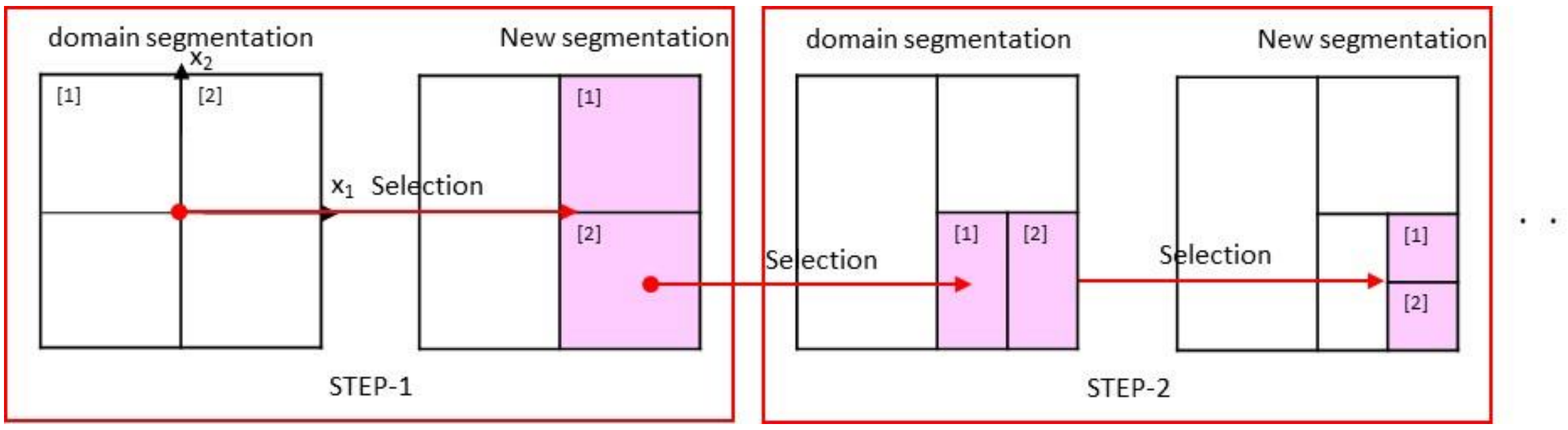

a) Modified segmentation

FIG.12: Modification of optimization using Monte Carlo method

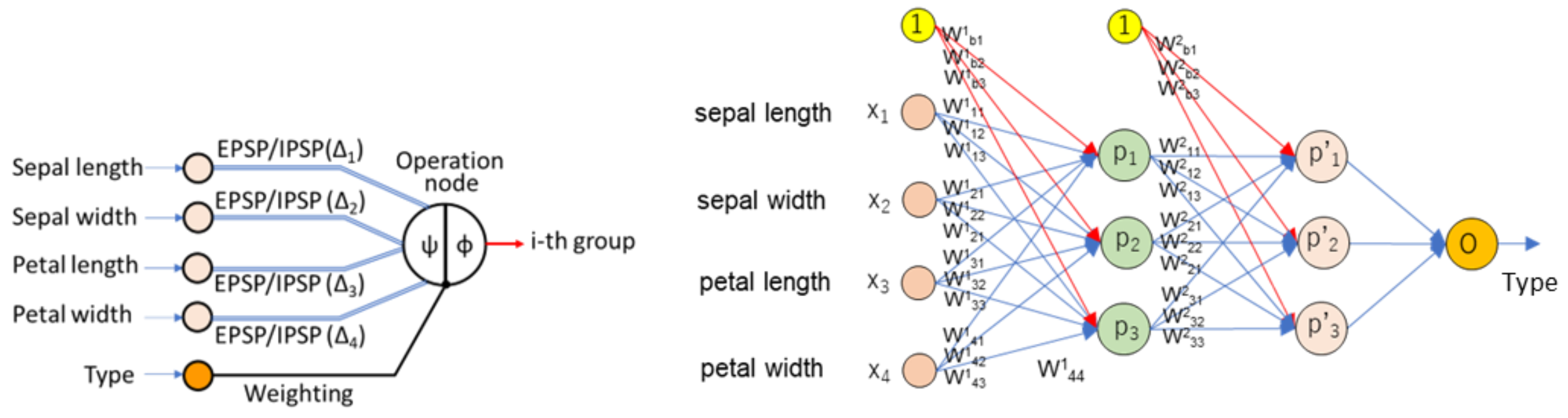

a) Configulation of SiNG b) Configuration of Neural Networks

FIG.13: Comparison between SiNG and Neural Networks of Iris Classification

Table.2: Example of Teaching signals

| DATA No | Sepal length (cm) | Sepal width (cm) | Petal length (cm) | Petal width (cm) | Flower type |
|---|---|---|---|---|---|
| 1 | 5.1 | 3.5 | 1.4 | 0.1 | 1 |
| 2 | 4.9 | 3 | 1.4 | 0.1 | 1 |
| 3 | 4.7 | 3.2 | 1.3 | 0.2 | 1 |
| 4 | 4.6 | 3.1 | 1.5 | 0.2 | 1 |
| : | : | : | : | : | : |
| 150 | 5.9 | 3 | 5.1 | 1.8 | 3 |

Table.3: Optimized weights on Neural Networks by MOST

| Weights | Results | Weights | Results | Weights | Results |
|---|---|---|---|---|---|
| $W^1_{11}$ | 1.19727 | $W^1_{41}$ | 0.95117 | $W^2_{21}$ | 1.75 |
| $W^1_{12}$ | -1.50391 | $W^1_{42}$ | -1.38770 | $W^2_{22}$ | 0.00098 |
| $W^1_{13}$ | -0.49902 | $W^1_{43}$ | -0.61328 | $W^2_{23}$ | 0.24707 |
| $W^1_{21}$ | 1.54395 | $W^1_{b1}$ | 1.88379 | $W^2_{11}$ | 0.25195 |
| $W^1_{22}$ | -1.02734 | $W^1_{b2}$ | 1.61328 | $W^2_{12}$ | 0.99609 |
| $W^1_{23}$ | -1.59277 | $W^1_{b3}$ | 1.53320 | $W^2_{13}$ | -0.12891 |
| $W^1_{31}$ | 0.83789 | $W^2_{11}$ | -0.50000 | $W^2_{b1}$ | -0.37500 |
| $W^1_{32}$ | 1.61035 | $W^2_{12}$ | -1.75000 | $W^2_{b2}$ | 1.81543 |
| $W^1_{33}$ | 1.63867 | $W^2_{13}$ | -2.00000 | $W^2_{b3}$ | -0.96680 |

Table.4: Optimized weights on SiNG

| $\Delta_1$ | $\Delta_2$ | $\Delta_3$ | $\Delta_4$ |
|---|---|---|---|
| 0.75 | 0.75 | 0.75 | 0.5 |

Table 5: Comparison between true Type and Prediction on TEST signals

| No. | Sepal_length | sepal_width | petal_length | petal_width | Type | Prediction | P(1) | P(2) | P(3) |
|---|---|---|---|---|---|---|---|---|---|
| 1 | 5 | 3.5 | 1.3 | 0.3 | 1 | 1 | 38 | 0 | 0 |
| 2 | 4.5 | 2.3 | 1.3 | 0.3 | 1 | 1 | 6 | 0 | 0 |
| 3 | 4.4 | 3.2 | 1.3 | 0.2 | 1 | 1 | 27 | 0 | 0 |
| 4 | 5 | 3.5 | 1.6 | 0.6 | 1 | 1 | 38 | 0 | 0 |
| 5 | 5.1 | 3.8 | 1.9 | 0.4 | 1 | 1 | 33 | 0 | 0 |
| 6 | 4.8 | 3 | 1.4 | 0.3 | 1 | 1 | 32 | 0 | 0 |
| 7 | 5.1 | 3.8 | 1.6 | 0.2 | 1 | 1 | 34 | 0 | 0 |
| 8 | 4.6 | 3.2 | 1.4 | 0.2 | 1 | 1 | 29 | 0 | 0 |
| 9 | 5.3 | 3.7 | 1.5 | 0.2 | 1 | 1 | 37 | 0 | 0 |
| 10 | 5 | 3.3 | 1.4 | 0.2 | 1 | 1 | 36 | 0 | 0 |
| 11 | 5.5 | 2.6 | 4.4 | 1.2 | 2 | 2 | 0 | 20 | 0 |
| 12 | 6.1 | 3 | 4.6 | 1.4 | 2 | 2 | 0 | 29 | 0 |
| 13 | 5.8 | 2.6 | 4 | 1.2 | 2 | 2 | 0 | 26 | 0 |
| 14 | 5 | 2.3 | 3.3 | 1 | 2 | 2 | 0 | 10 | 0 |
| 15 | 5.6 | 2.7 | 4.2 | 1.3 | 2 | 2 | 0 | 27 | 0 |
| 16 | 5.7 | 3 | 4.2 | 1.2 | 2 | 2 | 0 | 25 | 0 |
| 17 | 5.7 | 2.9 | 4.2 | 1.3 | 2 | 2 | 0 | 28 | 0 |
| 18 | 6.2 | 2.9 | 4.3 | 1.3 | 2 | 2 | 0 | 33 | 0 |
| 19 | 5.1 | 2.5 | 3 | 1.1 | 2 | 2 | 0 | 5 | 0 |
| 20 | 5.7 | 2.8 | 4.1 | 1.3 | 2 | 2 | 0 | 27 | 0 |
| 21 | 6.7 | 3.1 | 5.6 | 2.4 | 3 | 3 | 0 | 0 | 15 |
| 22 | 6.9 | 3.1 | 5.1 | 2.3 | 3 | 3 | 0 | 0 | 17 |
| 23 | 5.8 | 2.7 | 5.1 | 1.9 | 3 | 3 | 0 | 0 | 21 |
| 24 | 6.8 | 3.2 | 5.9 | 2.3 | 3 | 3 | 0 | 0 | 20 |
| 25 | 6.7 | 3.3 | 5.7 | 2.5 | 3 | 3 | 0 | 0 | 13 |
| 26 | 6.7 | 3 | 5.2 | 2.3 | 3 | 3 | 0 | 0 | 20 |
| 27 | 6.3 | 2.5 | 5 | 1.9 | 3 | 3 | 0 | 0 | 22 |
| 28 | 6.5 | 3 | 5.2 | 2 | 3 | 3 | 0 | 0 | 24 |
| 29 | 6.2 | 3.4 | 5.4 | 2.3 | 3 | 3 | 0 | 0 | 22 |
| 30 | 5.9 | 3 | 5.1 | 1.8 | 3 | 3 | 0 | 0 | 18 |

Table.6: Comparison of results

| Machine learning | Correct answer rate for Teaching signals | Correct answer rate for TEST signals |
|---|---|---|
| SiNG | 100% | 100% |
| Neural Networks with MOST | 99% | 93% |

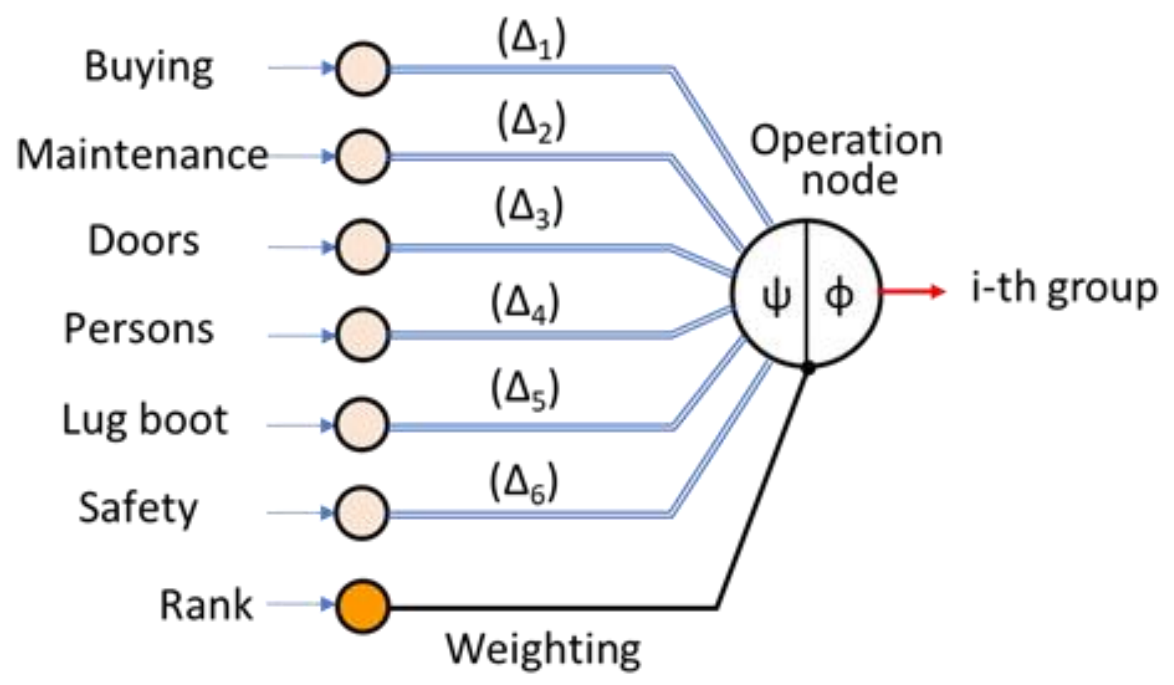

a) Configulation of SiNG

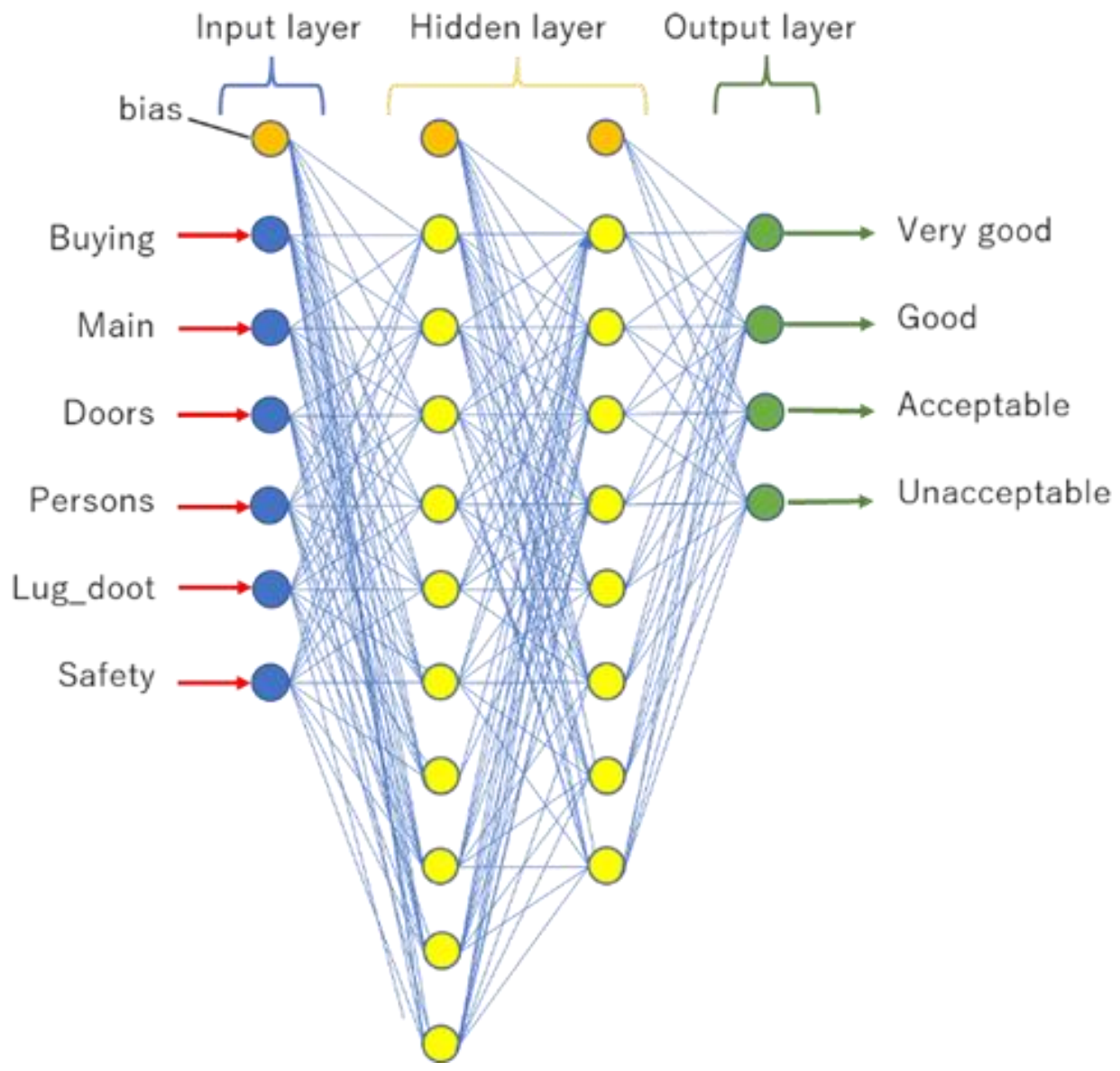

b) Configuration of Neural Networks

FIG.14: Comparison between SiNG and Neural Networks of Car Ranking Classification

Table.7: Example of Teaching signals

| No | Buying Price | Price of Maintenance | Doors | Persons Capacity | Lug_boot | Safety | Rank |
|---|---|---|---|---|---|---|---|
| 1 | 0 | 0 | 2 | 2 | 0 | 0 | 1 |
| 2 | 0 | 0 | 2 | 2 | 0 | 1 | 1 |
| 3 | 0 | 0 | 2 | 2 | 0 | 2 | 1 |
| 4 | 0 | 0 | 2 | 2 | 1 | 0 | 1 |
| 5 | 0 | 0 | 2 | 2 | 1 | 1 | 1 |
| ⋮ | ⋮ | ⋮ | ⋮ | ⋮ | ⋮ | ⋮ | ⋮ |
| 995 | 3 | 1 | 5 | 5 | 1 | 2 | 4 |

Table.8: Ranking of parameters

| Input data | 1 | 2 | 3 | 4 | 5 |
|---|---|---|---|---|---|
| Buying Price | Low | Medium | High | Very high | - |
| Price of Maintenance | Low | Medium | High | Very high | - |
| Doors | - | 2doors | 3doors | 4doors | 5more |
| Persons | - | 2people | - | 4people | More |
| Lug_boot | Small | Medium | Big | Small | - |
| Safety | Low | Medium | High | Low | - |

Table.9: Optimized weights on SiNG

| $\Delta_1$ | $\Delta_2$ | $\Delta_3$ | $\Delta_4$ | $\Delta_5$ | $\Delta_6$ |
|---|---|---|---|---|---|
| 2.55 | 0.1 | 2.55 | 0.1 | 0.1 | 0.1 |

Table.10: Comparison of results

| Machine learning | Correct answer rate for Teaching signals | Correct answer rate for TEST signals |
|---|---|---|
| SiNG | 100% | 94% |
| Neural Networks with MOST | 84% | 81% |

Table.11: Example of Teaching signals

| No. | Sex | Length | Diameter | Height | Whole weight | Shucked weight | Viscera weight | Shell weight | Rings |
|---|---|---|---|---|---|---|---|---|---|
| 1 | 1 | 0.455 | 0.365 | 0.095 | 0.514 | 0.2245 | 0.101 | 0.15 | 15 |
| 2 | 2 | 0.35 | 0.265 | 0.09 | 0.2255 | 0.0995 | 0.0485 | 0.07 | 7 |
| 3 | 1 | 0.53 | 0.42 | 0.135 | 0.677 | 0.2565 | 0.1415 | 0.21 | 9 |
| 4 | 3 | 0.44 | 0.365 | 0.125 | 0.516 | 0.2155 | 0.114 | 0.155 | 10 |
| 5 | 3 | 0.33 | 0.255 | 0.08 | 0.205 | 0.0895 | 0.0395 | 0.055 | 7 |
| ⋮ | ⋮ | ⋮ | ⋮ | ⋮ | ⋮ | ⋮ | ⋮ | ⋮ | ⋮ |
| 2000 | 3 | 0.35 | 0.25 | 0.07 | 0.1605 | 0.0715 | 0.0335 | 0.046 | 6 |

Table.12: Optimized weights on SiNG

| $\Delta_1$ | $\Delta_2$ | $\Delta_3$ | $\Delta_4$ | $\Delta_5$ | $\Delta_6$ | $\Delta_7$ | $\Delta_8$ |
|---|---|---|---|---|---|---|---|
| 1.125 | 0.25 | 0.25 | 0.1 | 0.375 | 0.375 | 0.375 | 0.375 |

Table.13: Comparison of results

| Machine learning | Correct answer rate for Teaching signals | Correct answer rate for TEST signals |
|---|---|---|
| SiNG | 100% | 86% |
| Egemen Sahin et al. | - | 79% |